\documentclass{article}

\usepackage{arxiv}

\usepackage[utf8]{inputenc} % allow utf-8 input
\usepackage[T1]{fontenc}    % use 8-bit T1 fonts
\usepackage{hyperref}       % hyperlinks
\usepackage{url}            % simple URL typesetting
\usepackage{booktabs}       % professional-quality tables
\usepackage{amsfonts}       % blackboard math symbols
\usepackage{nicefrac}       % compact symbols for 1/2, etc.
\usepackage{microtype}      % microtypography
\usepackage{lipsum}
\usepackage{svg}
\usepackage{graphicx}
\graphicspath{ {./images/} }
\usepackage{amsmath} 
\usepackage{pdflscape}
\usepackage{xcolor}
\usepackage[normalem]{ulem}
\usepackage{comment}
\usepackage{longtable}
\usepackage{tabularx}
\usepackage{caption} % for more complex captioning
\usepackage{subcaption}
\usepackage{longtable}
\usepackage{booktabs}
\usepackage[utf8]{inputenc}
\usepackage{changepage}

\usepackage{array}
\usepackage{float}
\newlength{\extralength}
\newlength{\fulllength}
\newcolumntype{C}{>{\centering\arraybackslash}X}

\title{From SAM to SAM 2: Exploring Improvements in Meta's Segment Anything Model}

\author{Athulya Sundaresan Geetha \textsuperscript{*} and Muhammad Hussain \\[1ex]
\begin{minipage}[t]{0.90\textwidth}
\centering
\scriptsize Department of Computer Science, Huddersfield University, Queensgate, Huddersfield HD1 3DH, UK; \\
\textsuperscript{*}Correspondence: U2282847@unimail.hud.ac.uk;
\end{minipage}}

\begin{document}

\maketitle
% Abstract (Do not insert blank lines, i.e. \\) 
\begin{abstract} The Segment Anything Model (SAM), introduced to the computer vision community by Meta in April 2023, is a groundbreaking tool that allows automated segmentation of objects in images based on prompts such as text, clicks, or bounding boxes. SAM excels in zero-shot performance, segmenting unseen objects without additional training, stimulated by a large dataset of over one billion image masks. SAM 2 expands this functionality to video, leveraging memory from preceding and subsequent frames to generate accurate segmentation across entire videos, enabling near real-time performance. This comparison shows how SAM has evolved to meet the growing need for precise and efficient segmentation in various applications. The study suggests that future advancements in models like SAM will be crucial for improving computer vision technology.
\end{abstract}

% Keywords

\section{Introduction}
In the rapidly advancing field of computer vision, target segmentation plays a vital role in enabling models to understand and interact with visual data. This process involves identification and isolation of objects within an image or video, a task essential for numerous applications, from autonomous vehicles and manufacturing ~\cite{zahid2023lightweight} to healthcare ~\cite{aydin2023domain} and renewable energy ~\cite{hussain2023review}. Conventional segmentation models often require extensive training on large datasets and fine-tuning to perform accurately across various contexts. However, the introduction of the Segment Anything Model (SAM), introduced by Meta in April 2023, marks a significant transformation in this domain.

SAM is designed to perform segmentation with minimal human intervention, providing robust zero-shot capabilities that enable it to segment objects it has never seen before. This innovation is underpinned by a pre-trained Vision Transformer, which serves as the image encoder. The image encoder processes visual data prior to any prompts being applied, transforming the raw image into a rich, high-dimensional representation. This step is fundamental in ensuring that the model can effectively interpret and respond to different types of prompts—whether they are text-based, clicks, or bounding boxes—leading to precise and efficient segmentation.

The efficiency of SAM had been further enhanced with the development of SAM2, which permeates the model’s capabilities to video segmentation. SAM 2 builds SAM 1, incorporating a memory mechanism that allows it to maintain consistency and accuracy across video frames by leveraging information from both past and future frames. This evolution not only manifests SAM's adaptability but also underscores the growing need for sophisticated models capable of handling complex visual data in real time.

This paper provides a detailed comparison of SAM and SAM2, highlighting how these models have been designed to meet the increasing demands of various applications. By examining the advancements in both variants, we emphasize the importance of continued innovation in computer vision technology for real-time image segmentation. Future advancements in models like SAM are likely to play a crucial role in expanding the boundaries of what is possible in the field of computer vision.

\section{Segment Anything Model}
Segment Anything Model (SAM) is a model used in computer vision for the segmentation of an image, created by Meta Research in April 2023, allowing users to generate masks by segmenting all objects in an image, objects related to the text prompt, or objects related to a single point on the frame. Its zero-shot performance can segment objects that have not been seen before without additional training, unlike other models that need fine-tuning, with over one billion image masks and 11 million images \cite{RN1, rath2023segment}.

With segmentation masks, the shape of the object is mapped, providing clarity about its dimensions and location. SAM enables fully automatic segmentation of an image with text-based input, relying less on training data and a lot of retraining processes.

SAM can segment areas based on the text input given by the users, namely clicking a single point, creating a bounding box, or marking a rough mask on an object. When clicking on an object in the image, SAM might mask different parts of the object if it is unsure about the primary object to be masked, thus producing multiple valid masks.

\subsubsection{Dataset}
The dataset includes 11 million images with high resolution and 1.1 billion segmentation masks with high quality generated by the data engine. The images were licensed by the provider working closely with the photographers. The average pixel size of the image was 3,300 × 4,950 but was downsampled by 1,500 pixels for accessibility purposes; despite downsampling, still, the images were still in high resolution. Faces and license plates are blurred in order to protect their identity and privacy.

\begin{figure}[H]
\begin{adjustwidth}{-\extralength}{0cm}
\centering
\includegraphics[height=8cm]{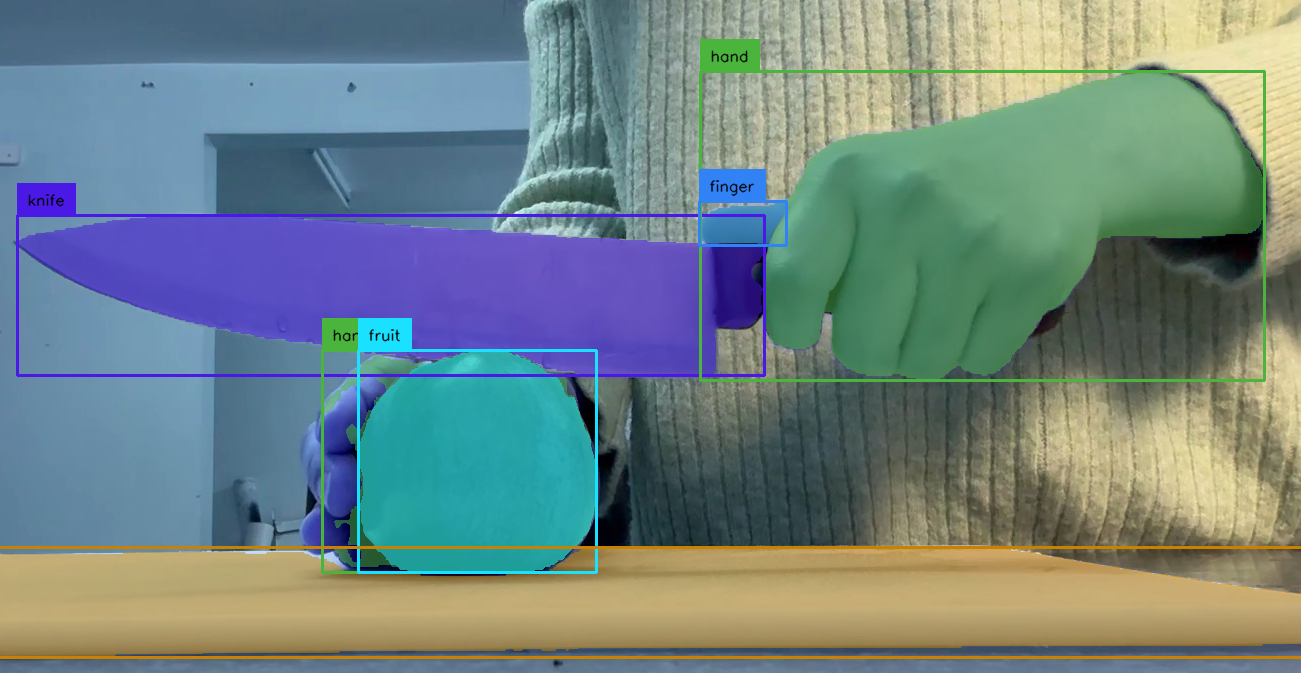}
\end{adjustwidth}
\caption{Segmentation example.}
\label{Figure:1}
\end{figure}

The dataset of 1.1 billion masks was assessed for quality, where 99.1\% of which are generated automatically. It is evident that 94\% of the masks have an IoU greater than 90\%, illustrating high quality. This dataset provides properties including size and concavity when compared with other datasets, making SA-1B an important source to develop computed vision models. An example of segmentation in the kitchen dataset is given in Figure \ref{Figure:1}.

\subsection{Components in SAM}
SAM for image segmentation consists of three components: an image encoder, a prompt encoder, and a mask decoder.

\subsubsection{Image Encoder}
The image encoder in SAM is responsible for processing images before the model's segmentation prompts are applied. It employs a pre-trained Vision Transformer \cite{RN2} to perform this initial encoding. This Vision Transformer, trained on large image data, extracts and represents complex visual features from the image, such as edges, textures, shapes, and objects. By transforming the image to a high-dimensional feature space, the encoder prepares an informative input for the subsequent segmentation steps. This preprocessing enhances the model’s ability to accurately and efficiently generate segmentation masks based on various prompts, whether text-based, click-based, or bounding boxes. The use of a pre-trained Vision Transformer ensures that the encoder captures a broad range of visual patterns and details, facilitating robust and precise object segmentation.

\subsubsection{Prompt Encoder}
The prompt encoder handles the following prompts: sparse, namely points, boxes, and texts, and dense, such as masks, using positional encodings \cite{RN3} along with learning embeddings and a CLIP text encoder \cite{RN4}, processing coordinates with front and background information, bounding box points, and tokens for text.

\subsubsection{Mask Decoder}
Utilizing a modified Transformer decoder block  \cite{RN5}, the mask decoder converts image and text input into masks, incorporating self-attention and cross-attention, followed by upsampling and a dynamic classifier \cite{RN6}.

\subsection{Data Engine}
The data engine and training of the SAM have the following three annotation stages: manual annotation, semi-automatic, and fully automatic annotation.

\subsubsection{Manual Annotation}
During the model-assisted manual annotation, 120,000 images were segmented using pre-trained SAM model, generating 4.3 million masks, on which SAM was retrained. The users clicked on the front and background objects, refining masks with brush and eraser tools. While focusing on objects that the annotators could identify, they labeled identifiable objects and proceeded to the next image if a mask took over 30 seconds to annotate \cite{RN7}. 

\subsubsection{Semi-automatic Annotation}
Secondly, the users added 5.9 million masks for 180,000 images, retraining model to improve segmentation. Confident masks were auto-detected, helping annotators. The average time for annotation of a mask was 34 seconds, with masks in an image increasing from 44 to 72, Total masks reached 10.2 million.

\subsubsection{Fully-automatic Annotation}
Finally, in the fully automatic annotation process, SAM trained on over 10 million masks by annotations for 11 million images, resulting in 1.1 billion masks. Improvements include different masks and the model's awareness of unidentifiable image detecting valid masks from a 32 × 32 grid. Confident masks were determined using IoU prediction and non-maximal suppression.

\subsection{Architecture of SAM}
The process starts with an image as an input which is passed through the image encoder to generate image embeddings, representing the image that a model can understand. The prompt encoder processes various inputs such as points, bounding boxes, and text prompts, thereby guiding the model to focus on the objects in the image. The prompt data along with image encoder is fed into a lightweight mask decoder. This mask decoder generates multiple segmentation masks corresponding to the different images.

\begin{figure}[H]
\begin{adjustwidth}{-\extralength}{0cm}
\centering
\includegraphics[width=15cm]{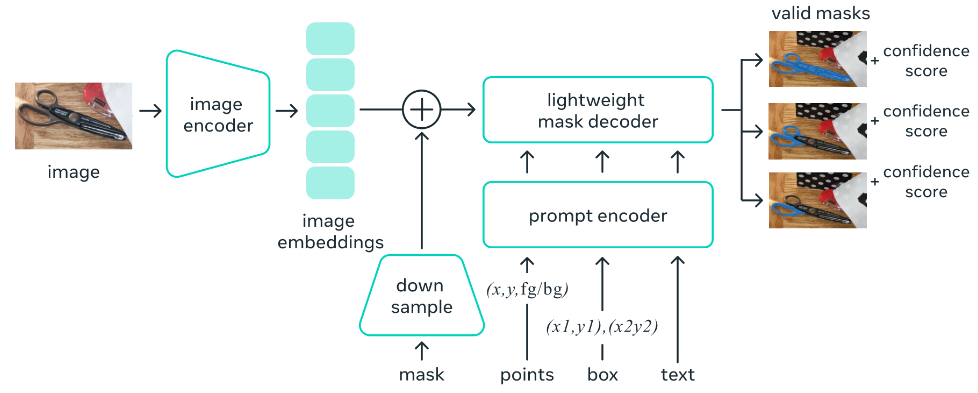}
\end{adjustwidth}
\caption{Architecture of Segment Anything Model \cite{rath2023segment}}
\label{Figure:2}
\begin{minipage}{15cm} % Adjust width to match the figure width
\vspace{0.5em} % Space between caption and footnote
\footnotesize
\end{minipage}
\end{figure}

Each mask is associated with the confidence score, providing how confident the model is in the particular object segmentation. In final, the output consists of the most valid masks, whereby each mask is annotated with its confidence score (Figure \ref{Figure:2}).

\subsection{Applications}
SAM enhances augmented reality experiences by detecting objects, helping users with reminders, and allowing lifting of objects in the virtual realm. It segments medical images and cell microscopy with no retraining. By integrating it with diffusion models, the creation of masks was automated, thereby making it easier to fill in missing parts in an image by zero-shot labeling. Additionally, it enhances the quality of object segmentation by generating accurate masks for the dataset. It also assisted with pre-trained models to recommend annotations for labeling an image. Synthetic generation of images uses SAM for creating diverse labeled datasets \cite{RN7, RN8}.

\subsection{Limitations}
SAM performed well in the segmentation of an image but not well with video data. To segment video frames, it needs to be combined with other deep learning models like object detection, which could cause delays. This makes SAM less effective for real-time processing.

\section{Segment Anything Model 2}
SAM2 extends the functionality of SAM to both video and image. For every frame of the video, SAM2 identifies the spatial boundaries of object using point, box, and prompts. For single frame, it functions like SAM, employing a prompt-based mask decoder. This decoder takes in a frame embedding with the prompts to generate a segmentation mask, which can be refined by adding more prompts continuously \cite{RN9, RN10, RN11, RN12}.

SAM2 does not generate embeddings directly from an image encoder; rather, it makes use of a combination of earlier forecasts and prompts from the previous and future frames in relation to the current one. A memory encoder creates these memories by processing continuous predictions and storing them in a memory bank. Memory attention refines embedding with the data stored for the mask decoder.

\begin{figure}[H]
\begin{adjustwidth}{-\extralength}{0cm}
\centering
\includegraphics[width=15cm]{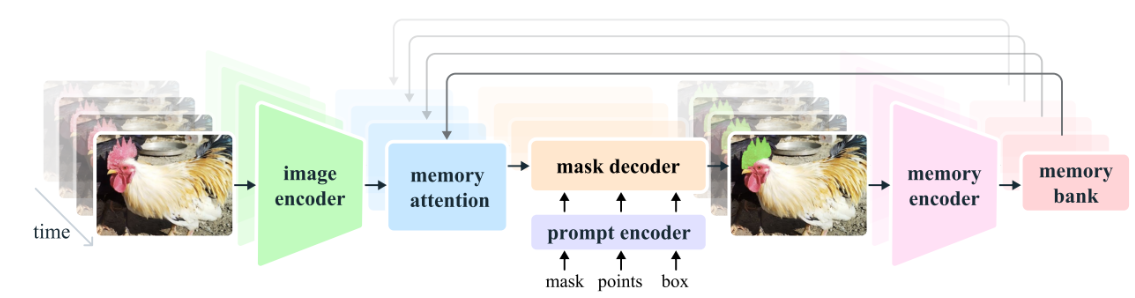}
\end{adjustwidth}
\caption{Architecture of Segment Anything Model 2 \cite{RN9}}
\label{Figure:3}
\begin{minipage}{15cm} % Adjust width to match the figure width
\vspace{0.5em} % Space between caption and footnote
\footnotesize
\end{minipage}
\end{figure}

The model quickly generates a valid segmentation mask for the frame after recognizing a variety of prompts on any video frame. Upon receiving prompts on one or more frames, the model uses this data to generate a segmentation mask for the whole video, with other prompts enabling continuous improvement. Both online and offline settings are used to assess the model, covering interactive video segmentation with annotations over several frames (Figure \ref{Figure:3}).

\subsection{Dataset}
The SA-V dataset is a resource with 50.9K videos and 642.6K masklets, making it 53 times larger than the largest video object segmentation (VOS) dataset. With an average duration of 14 seconds, the videos comprise 54\% indoor and 46\% outdoor scenes. The 47-country dataset involves a variety of real-world scenarios that cover day-to-day scenarios. Moreover, the research on VOS can benefit from this large collection of videos and annotations.

With the combination of 451.7K automatically generated masklets and 190.9K manually annotated masklets, SA-V dataset has an advantage in mask quantity against other datasets. High relevance and quality are provided by the annotation images, focusing on difficult tasks like dynamic movement of objects and occluded images. This dataset is further improved by internal data augmentation, which adds 69.6K masklets and 62.9K videos for training.

\subsection{Architecture of SAM 2}

\subsubsection{Image Encoder}

\begin{figure}[H]
\begin{adjustwidth}{-\extralength}{0cm}
\centering
\includegraphics[width=15cm]{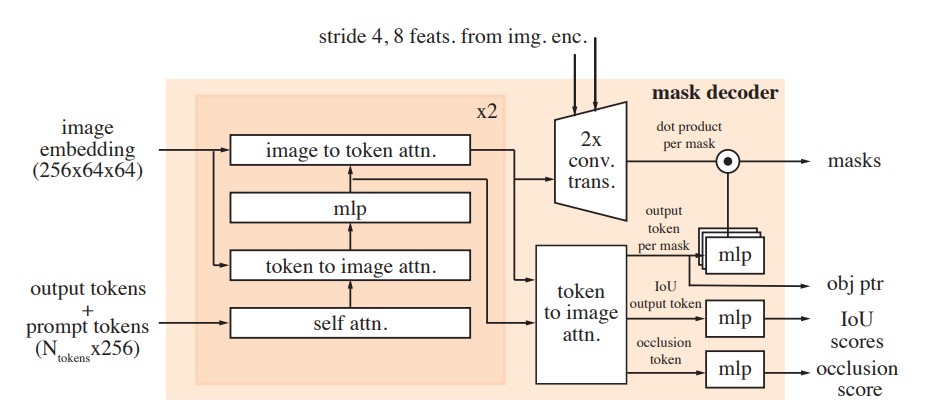}
\end{adjustwidth}
\caption{Image encoding \cite{RN10}}
\label{Figure:4}
\begin{minipage}{15cm} % Adjust width to match the figure width
\vspace{0.5em} % Space between caption and footnote
\footnotesize
\end{minipage}
\end{figure}

The image encoder, a layered masked encoder (Hiera), processes video frames one by one, utilizing multiscale features during decoding. To generate the image embeddings for each frame, a feature pyramid network is implemented to fuse the stride 16 and 32 features from stages 3 and 4 of the Hiera encoder, respectively \cite{RN13}. However, the memory attention mechanism does not include the stride 4 and 8 features from stages 1 and 2, respectively, but incorporates them into the mask decoder's upsampling layers (Figure \ref{Figure:4}), producing high-resolution segmentation details. In accordance with another paper, windowed absolute positional embeddings are employed \cite{RN14}. RPB provided positional data spanning windows in the encoder of the image; however, a simple method includes interpolating global positional  \cite{RN15} embedding to cover the windows.

\subsubsection{Memory Attention}
Memory attention aligns current frame features with past data such as features of frames and predictions and prompts, arranging transformer blocks where the first block extracts features. This mechanism uses four layers and utilize 2D spatial Rotary Positional Embedding in addition to sinusoidal absolute positional embedding in self-attention and cross-sectional layers, with tokens to point objects excluded \cite{RN16, RN17}.

\subsubsection{Prompt Encoder and Mask Decoder}
The prompt encoder, like SAM, processes various prompt types. Sparse prompts utilize positional encoding, whereas dense prompts, namely masks, use convolutional layers. The mask decoder receives encoded prompts and conditioned frames as inputs, along with a skip connection from the image encoder, enhancing high-resolution detail.

\subsubsection{Memory Encoder and Memory Bank}
Predictions and embeddings from the image encoder for the frame are processed by the memory encoder and saved for future use. It holds the past predictions for the object and keeps track of prompts from the prompt encoder for relevant frames.

\subsection{Data Engine}

\subsubsection{SAM}

In Phase 1, SAM was employed to generate initial masks for video frames extracted at 6 frames per second, assisting annotators. They then refined these masks using pixel-precise tools namely brushes and erasers, without temporal tracking models. Each frame was individually annotated, slowing the process with an average time of 37.8 seconds per frame. Although it is a time-consuming process, this method obtained spatial annotations of the frames in high quality, producing 16K masklets with 1.4K videos.

\subsubsection{SAM and SAM 2}

In Phase 2, SAM and SAM 2 collaborated with human annotators. SAM 2 was tasked to create spatio-temporal masklets by propagating masks from SAM 1 and input given by annotators to process video frames. Spatial masks were created on the first frame using SAM 1 and extended across subsequent frames using SAM 2 mask. Using SAM and tools, the users could fine-tune the masks at any frame point and then reapply SAM 2 mask for updates. SAM 2 mask was retrained again using new data after being trained SAM 2 on Phase 1 data and other datasets. This method increased the speed by taking 63.5K masklets and cutting down on annotation time to 7.4 seconds per frame. Memory limitations made manual annotations necessary for intermediate frames; however, this phase established the background for SAM 2, integrating interactive segmentation and mask propagation into a single model.

\subsubsection{SAM 2}

During Phase 3, SAM 2, which responds to all prompts and uses temporal memory to track objects, was used with little aid from the annotators and needed only minor adjustments for betterment, generating 197K masklets. At 4.5 seconds per frame, there seems to be a notable decrease in annotation time. Not only the annotation time was decreased but also efficiency was increased by retraining the SAM  2 five times.

\subsection{Procedure}
\begin{figure}[H]
\begin{adjustwidth}{-\extralength}{0cm}
\centering
\includegraphics[width=15cm]{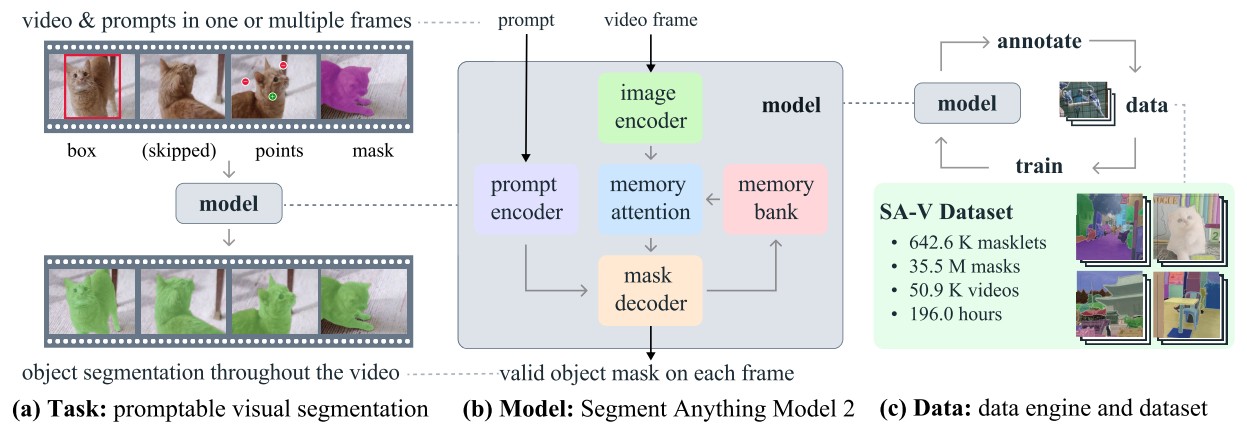}
\end{adjustwidth}
\caption{Promptable Visual Segmentation \cite{RN10}}
\label{Figure:5}
\begin{minipage}{15cm} % Adjust width to match the figure width
\vspace{0.5em} % Space between caption and footnote
\footnotesize
\end{minipage}
\end{figure}

Figure \ref{Figure:5} represents a system for promptable visual segmentation employing SAM 2. This model processes videos by taking into consideration prompts, including bounding boxes, points, and masks, to identify objects in one or multiple frames and generates segmentation masks throughout the whole video. To maintain consistency across all the frames, the models look at the previous frames, ensuring that the object is correctly segmented in the entire video. By training with the SA-V dataset, which comprises 35.5M masks, 50.9K videos, 642.6K masklets, and 196-hour footage, the model can perform well.

\subsection{Applications}
Combining SAM2 with generative video models provides new possibilities for editing the videos and produces unique visual effects. SAM 2 supports segmentation and improves the accuracy of anatomical structure identification in the fields of imaging in the medical sector and research. SAM 2 also enhances awareness in autonomous vehicles, facilitating better navigation and elimination of obstacles. It also speeds up the process by creating annotated datasets, reducing time and effort when compared with manual annotations.

\subsection{Limitations}
While SAM 2 is extremely effective in static image and video, segmenting objects in overcrowded place, after prolonged occlusions, or when the shot changes can be difficult task for SAM 2, which might result in confusion or worse loss of tracking the object. SAM 2 addresses this by allowing adding prompts to any frames, where refinement clicks can be used to quickly recover the model if it gets lost. It struggles to distinguish between objects that look similar and to track objects with minute details, particularly while moving quickly. In these scenarios, improving the model with more precise motion modeling might aid in lowering the mistakes.

SAM 2 can track multiple objects simultaneously, but it handles them separately, depending on shared per-frame embeddings and ignoring inter-object communication, using shared object-level context to increase productivity. Currently, to ensure masklet quality and correct frames, the data engine relies on manual annotations. However, in the future, this can be automated to increase efficiency.

\section{Comparison of SAM and SAM 2}

SAM is a segmentation model to identify and segment any object in an image using various input methods such as points, bounding boxes, and text prompts, offering a wide range of segmentation capabilities for various tasks. To create high-quality segmentation masks, SAM utilizes a transformer-based architecture with attention mechanisms, supporting various input types. The tool's user-friendliness lies in its ability to allow users to interactively refine segmentations by adding prompts if initial results are not satisfactory, resulting in high-quality outcomes.

SAM 2 is expected to improve SAM's fundamental capabilities with increased precision, effectiveness, and new features. Moreover, further developments, such as more advanced attention mechanisms, better handling, and better management of various input types, are considered. It is expected to provide enhancements to improve the user's experience, simplifying the segmentation process, thereby making it smoother and faster. A comparison between the variants is presented in \ref{tab:sam_comparison}.

SAM may require significant computational resources in processing high-resolution images, while SAM 2 is expected to be optimized for better computational efficiency by reducing hardware requirements and improving computational speed. Overall, SAM excels in performance and adaptability, whereas SAM 2 is projected to enhance accuracy, user interaction, and efficiency with new capabilities.

\begin{table}[H]
\caption{Comparison of SAM and SAM2 key features.\label{tab:sam_comparison}}
    \begin{adjustwidth}{-\extralength}{0cm}
        \begin{tabularx}{\fulllength}{CCCCC}
            \toprule
            \textbf{Feature} & \textbf{SAM (Segment Anything Model)} & \textbf{SAM2} \\
            \midrule
            Objective & General-purpose image segmentation & Likely an improved version with enhanced features \\
            \midrule
            User Interaction & Interactive refinement through points, bounding boxes, text prompts & Potentially smoother interaction with new controls and better handling of ambiguous inputs \\
            \midrule
            Applications & Image editing, medical imaging, general object segmentation & Similar applications with potential for new use cases \\
            \midrule
            Computational Efficiency & Requires substantial resources, especially for high-res images & Likely optimized for better efficiency, reducing hardware requirements \\
            \midrule
            Capabilities & Primarily focused on still images; limited support for video segmentation & Likely enhanced video capabilities, including improved tracking and segmentation over time \\
            \midrule
            Image Encoder & Utilizes a pre-trained Vision Transformer to extract and represent complex visual features from images & Employs a layered masked encoder (Hiera) with a feature pyramid network to fuse multiscale features, optimizing high-resolution detail \\
            \midrule
            Prompt Encoder & Handles sparse prompts (points, boxes) and dense prompts (masks) using positional encodings and a CLIP text encoder & Similar handling of prompts with enhancements, potentially incorporating more advanced embeddings and better integration of spatial and textual data \\
            \midrule
            Mask Decoder & Uses a modified Transformer decoder block with self-attention and cross-attention for generating masks & Optimized mask decoder with upsampling layers and improved resolution handling, leveraging additional memory and attention mechanisms \\
            \midrule
            Data Engine & Manual Annotation: 120,000 images segmented using SAM, generating 4.3 million masks. Semi-Automatic Annotation: Added 5.9 million masks for 180,000 images. Fully-Automatic Annotation: SAM trained on over 10 million masks, resulting in 1.1 billion masks. & Phase 1: Initial masks for video frames at 6 fps, refined manually with high spatial quality but slow annotation. Phase 2: Spatio-temporal masklets using SAM 1 and SAM 2, significantly improving speed and efficiency. Phase 3: SAM 2 with minimal manual adjustments, producing 197K masklets with reduced annotation time and increased efficiency. \\
            \midrule
            Limitations & May struggle with very fine details or highly complex scenes; computationally intensive & Potentially mitigates some of SAM's limitations but may still face challenges with very high complexity or extremely large-scale scenarios \\
            \bottomrule
        \end{tabularx}
    \end{adjustwidth}
\end{table}

SAM represents a significant advancement in image segmentation. It handles over 1.1 billion image masks and 11 million images. Its zero-shot capabilities allow to generate accurate masks for object segmentation without previous training. This is done by a pre-trained Vision Transformer for image encoding, a prompt encoder for various input types, and a modified transformer decoder for mask generation.

The accuracy of the created masks is emphasized by the fact that 94\% of the masks have an Intersection over Union score larger than 90\%, demonstrating the quality of the data. Moreover, the annotation processes ensure that SAM produces more reliable and high-quality results.

However, the majority of SAM's abilities are focused on segmenting still images, where its efficiency and accuracy are better when compared with video segmentation. 

SAM 2 extends its functionality to video segmentation. It has a memory encoder that improves segmentation accuracy across video sequences utilizing past and future frames. SAM's temporal memory mechanism improves upon SAM's static image-focused approach by providing more coherent and accurate masks. The architecture of SAM 2 introduces several improvements, involving a layered masked encoder, Hiera, and memory attention mechanism. These, in turn, facilitate the processing of high-resolution features and the integration of temporal data, improving the segmentation process in videos. Additionally, the SA-V dataset used for training underscores its capabilities with its large collection of video frames and masklets.

Despite the advantages, SAM 2 has limitations, where it cannot detect objects well when the scenes are cluttered and/or if the occlusion is prolonged for a long time. In addition, its handling of multiple objects is limited as it processes each object independently.

\section{Conclusions}
Overall SAM can be characterised as a powerful tool for image segmentation, enabling developers to identify and segment objects in images with various input methods like points, bounding boxes, and text prompts. Its transformer-based architecture and attention mechanisms assists with the creation of high-quality segmentation masks, and its interactive features let users refine these masks for better accuracy. However, SAM can be resource-intensive, especially when working with high-resolution images.

SAM2, the next version, is expected to improve on SAM's strengths by presenting higher precision, efficiency, and user interaction. SAM2’s optimised image encoder and mask decoder, along with improved handling of prompts and reduced hardware needs, are expected to make it even more effective at handling complex and large-scale tasks.

In summary, while SAM is already a highly adaptable and effective model for image segmentation, SAM2 is anticipated to make the process smoother, faster, and more efficient. These improvements highlight the ongoing need for innovation in computer vision, ensuring that future models can keep up with the growing demand for accuracy and efficiency in various applications.

%%%%%%%%%%%%%%%%%%%%%%%%%%%%%%%%%%%%%%%%%%
\begin{adjustwidth}{-\extralength}{0cm}

\bibliographystyle{unsrt}  % Changes bibliography style to unsorted
\bibliography{ref}  % This points to the filename of your BibTeX file without the .bib extension

\end{adjustwidth}
\end{document}